\documentclass{article}

\usepackage{arxiv}

\usepackage[utf8]{inputenc} 
\usepackage[T1]{fontenc}    
\usepackage{hyperref}       
\usepackage{url}            
\usepackage{booktabs}       
\usepackage{amsfonts}       
\usepackage{nicefrac}       
\usepackage{microtype}      
\usepackage{lipsum}
\usepackage{graphicx}
\usepackage{amsmath} 
\usepackage{multirow}
\usepackage{url,hyperref,lineno,microtype,subcaption}
\usepackage{caption}
\usepackage{graphicx}
\usepackage{comment}
\usepackage{booktabs}
\graphicspath{ {./images/} }

\title{Persistent Homology for MCI Classification: A Comparative Analysis between Graph and Vietoris-Rips Filtrations}

\author{
 Debanjali Bhattacharya \\
 Centre for Brain Research\\
 Indian Institute of Science\\
 Bengaluru, India \\
  \texttt{debanjali@cbr-iisc.ac.in} \\
  \And
 Rajneet Kaur \\
 Centre for Brain Research\\
 Indian Institute of Science\\
 Bengaluru, India \\
 \texttt{Rajneetkaur2620@gmail.com} \\
  \And
 Ninad Aithal \\
 Vision and AI Lab\\
 Indian Institute of Science\\
 Bengaluru, India \\
 \texttt{ninadaithal@iisc.ac.in} \\
  \And
  Neelam Sinha \\
 Centre for Brain Research\\
 Indian Institute of Science\\
 Bengaluru, India \\
 \texttt{neelam@cbr-iisc.ac.in} \\
  \And
  Thomas Gregor Issac \\
 Centre for Brain Research\\
 Indian Institute of Science\\
 Bengaluru, Karnataka, India \\
 \texttt{thomasgregor@cbr-iisc.ac.in} \\
 }

\begin{document}
\maketitle
\begin{abstract}
Mild cognitive impairment (MCI), often linked to early neurodegeneration, is characterized by subtle cognitive declines and disruptions in brain connectivity. The present study offers a detailed analysis of topological changes associated with MCI, focusing on two subtypes: Early MCI and Late MCI. This analysis utilizes fMRI time series data from two distinct populations: the publicly available ADNI dataset (Western cohort) and the in-house TLSA dataset (Indian Urban cohort). Persistent Homology, a topological data analysis method, is employed with two distinct filtration techniques—Vietoris-Rips and graph filtration—for classifying MCI subtypes. For Vietoris-Rips filtration, inter-ROI Wasserstein distance matrices between persistent diagrams are used for classification, while graph filtration relies on the top ten most persistent homology features. Comparative analysis shows that the Vietoris-Rips filtration significantly outperforms graph filtration, capturing subtle variations in brain connectivity with greater accuracy. The Vietoris-Rips filtration method achieved the highest classification accuracy of 85.7\% for distinguishing between age and gender matched healthy controls and MCI, whereas graph filtration reached a maximum accuracy of 71.4\% for the same task. This superior performance highlights the sensitivity of Vietoris-Rips filtration in detecting intricate topological features associated with neurodegeneration. The findings underscore the potential of persistent homology, particularly when combined with the Wasserstein distance, as a powerful tool for early diagnosis and precise classification of cognitive impairments, offering valuable insights into brain connectivity changes in MCI.
\end{abstract}


\section{Introduction}
Mild cognitive impairment (MCI) has become a prominent subject of investigation and clinical attention as it represents an intermediate stage between the anticipated cognitive decline associated with normal aging and the more severe cognitive and functional deficits observed in dementia (\cite{mosti_2019_differentiating, petersen_2008_mild}). MCI is characterized by a measurable decline in cognitive abilities, such as memory, language, or executive function, that is greater than expected for an individual's age and education level, yet does not significantly interfere with their ability to perform everyday activities. Although complex functional tasks, such as preparing meals, or shopping, may take more time or be performed less efficiently than before (\cite{gauthier_2006_mild, knopman_2014_mild}). It is prevalent in older adults. While some individuals with MCI remain stable or even return to normal cognitive function over time, over half of them progress to dementia within five years (\cite{gauthier_2006_mild}).Therefore, MCI can be viewed as a potential precursor to dementia. MCI is typically assessed by measuring impairment in standard memory or other cognitive tests. Specifically, it is evaluated in terms of a decline in episodic memory performance that falls below the norm adjusted for age and education. In order to gain a better understanding of the progression of mild cognitive impairment (MCI), the Alzheimer's Disease Neuroimaging Initiative (ADNI) has divided MCI into two stages: early MCI (EMCI) and late MCI (LMCI). EMCI is characterized by cognitive impairment that falls between 1 and 1.5 SD below the average on WMS-R Logical Memory II Story A test, while LMCI is characterized by performance that falls 1.5 SD below the average (\cite{aisen_2010_clinical}). 
According to the ADNI, both individuals with LMCI and EMCI were found to have an increased risk of AD-associated dementia. The annual conversion rate from MCI to dementia was 17.5\% for subjects with LMCI and 2.3\% for those with EMCI (\cite{weiner_2015_2014}). The diagnosis of mild cognitive impairment and its sub-types thus is of critical importance, as it can help identify individuals at an elevated risk of progressing to Alzheimer's disease or other forms of dementia. The identification of diagnostic markers that are highly sensitive and develop as the disease progresses can greatly assist physicians in making precise diagnoses. In recent years, there have been notable advancements in brain imaging, particularly with fMRI, which provide valuable insights into the impact of MCI on brain function. By analyzing changes in different brain networks, we can enhance our understanding of how MCI disrupts communication within the brain. These disruptions are crucial in detecting the disease at an early stage, potentially allowing for interventions that can slow down or reverse cognitive decline associated with MCI and its sub-types.

In the present study we employ persistent homology, an advanced tool in computational topology, to investigate potential variations in topology among different sub-types of MCI and compare them to a healthy control group (HC). 
Persistent homology is a powerful approach within the field of algebraic topology that falls under the umbrella of topological data analysis. It offers a robust framework for examining the topological properties of data, particularly in relation to shape and structure. The objective of persistent homology is to trace how topological features on a given space emerge and vanish as the scale value gradually changes. This process, known as filtration, allows us to understand how these features remain consistent or evolve when observing the spaces at different levels of detail (\cite{edelsbrunner_2017_persistent}). Persistent homology has gained significant traction in recent years, particularly in the analysis of fMRI data (\cite{abdallah_2022_statistical, catanzaro_2023_topological}). This approach is highly valued for its ability to account for non-neural variability (\cite{kumar_2023_the}) and its extensive use in medical research. By capturing the underlying structure and relationships within complex medical data, this innovative method holds great promise for generating new insights and improving the accuracy of clinical decision-making. Notably, its application has been observed in the analysis of Autism Spectrum Disorder (\cite{alirezataleshjafadideh_2022_topological}), Schizophrenia (\cite{stolz_2021_topological}), and brain tumor analysis (\cite{bhattacharya2024analyzingbraintumorconnectomics}). Moreover, it has been utilized to examine differences in visual brain networks (\cite{bhattacharya2023imagecomplexitybasedfmribold}).

The current study is an extension of our previous work \cite{aithal2024leveragingpersistenthomologydifferential}, where we leveraged persistent homology using Vietoris-Rips filtration for the differential diagnosis of Mild Cognitive Impairment (MCI). In this work, we aim to rigorously compare the effectiveness of graph filtration with the previously utilized Vietoris-Rips filtration. In the previous study Wasserstein distance matrices computed from persistent homology at different dimensions are used as feature for classification. Contrary to this, in the present study we hypothesize that persistent homology, particularly when utilizing raw persistence features generated from graph filtration can effectively differentiate between healthy individuals and those at various stages of MCI, including Early and Late MCI.
However, our findings reveal that while graph filtration offers valuable insights into the topological structure of brain connectivity, the classification results using Vietoris-Rips filtration are consistently superior. The Vietoris-Rips approach not only surpasses graph filtration in distinguishing MCI subtypes but also outperforms many state-of-the-art methods in MCI classification. These results highlight the greater efficacy of Vietoris-Rips filtration in extracting meaningful topological features, making it a more robust tool for clinical applications in early MCI detection and diagnosis. 
This findings of this study provide crucial insights into the comparative strengths of these two filtration approaches. Vietoris-Rips filtration proves to be more robust in capturing the intricate topological patterns in brain connectivity that are essential for accurate diagnosis. This superiority reinforces its potential for use in clinical settings, offering a more reliable framework for early detection and classification of MCI subtypes. Additionally, this research highlights the importance of choosing the appropriate filtration method in persistent homology applications, contributing new knowledge to the field of topological data analysis for disease classification. Our work serves as a guide for future studies, emphasizing that the choice of filtration has a significant impact on the outcomes of persistent homology-based classification.

\section{Materials and Methods}

\subsection{Dataset Description}

The study employs fMRI images from two distinct population cohorts. The baseline analysis utilizes subjects from the publicly available Alzheimer’s Disease Neuroimaging Initiative (ADNI) dataset (\cite{jack_2008_the}), characterized by a repetition time (TR) of 3000 ms and an echo time (TE) of 30 ms. This is complemented by our in-house cohort from the TATA Longitudinal Study for Aging (TLSA), which features a TR of 3200 ms and a TE of 30 ms. The TLSA cohort, an urban study, is dedicated to the long-term investigation of risk and protective factors associated with dementia in India. Both cohorts used sagittal plane imaging with 3D acquisition. Participants diagnosed with MCI had no underlying neurodegenerative diseases besides MCI itself, while healthy subjects had no previous instances of cognitive impairment, stroke, or significant psychiatric disorders.
The ADNI cohort includes the following groups: EMCI ($N = 162$, M:F = $59:103$, Age: $72.3\pm6.7$), LMCI ($N = 141$, M:F = $86:55$, Age: $72.1\pm7.8$), and HC ($N = 177$, M:F = $81:96$, Age: $75.1\pm6.3$). The efficacy of our proposed methodology is further evaluated using our in-house TLSA cohort, which consists of gender and age-matched MCI ($N = 35$, M:F = $20:15$, Age: $63.9\pm9.04$) and HC ($N = 35$, M:F = $20:15$, Age: $63.8\pm9.1$) subjects. In our in-house MCI cohort, individuals were diagnosed with MCI according to specific criteria, including a Clinical Dementia Rating (CDR) score of 0.5, which serves as the current gold standard for assessing the stages of dementia.
The study utilizes fMRI time series from Dosenbach Regions of Interest (ROIs), encompassing a total of 160 ROIs selected from six classical brain networks. All fMRI images were processed using a consistent preprocessing pipeline, which included motion correction, slice timing adjustment, normalization to the standard MNI space, and regression to account for nuisance variables. These preprocessing steps were conducted using FSL (FMRIB Software Library) version 6.0.6 \cite{jenkinson_2012_fsl}.

\subsection{Overview on Proposed Methodology}

\begin{figure}[h]
    \centering
    \includegraphics[width=\textwidth]{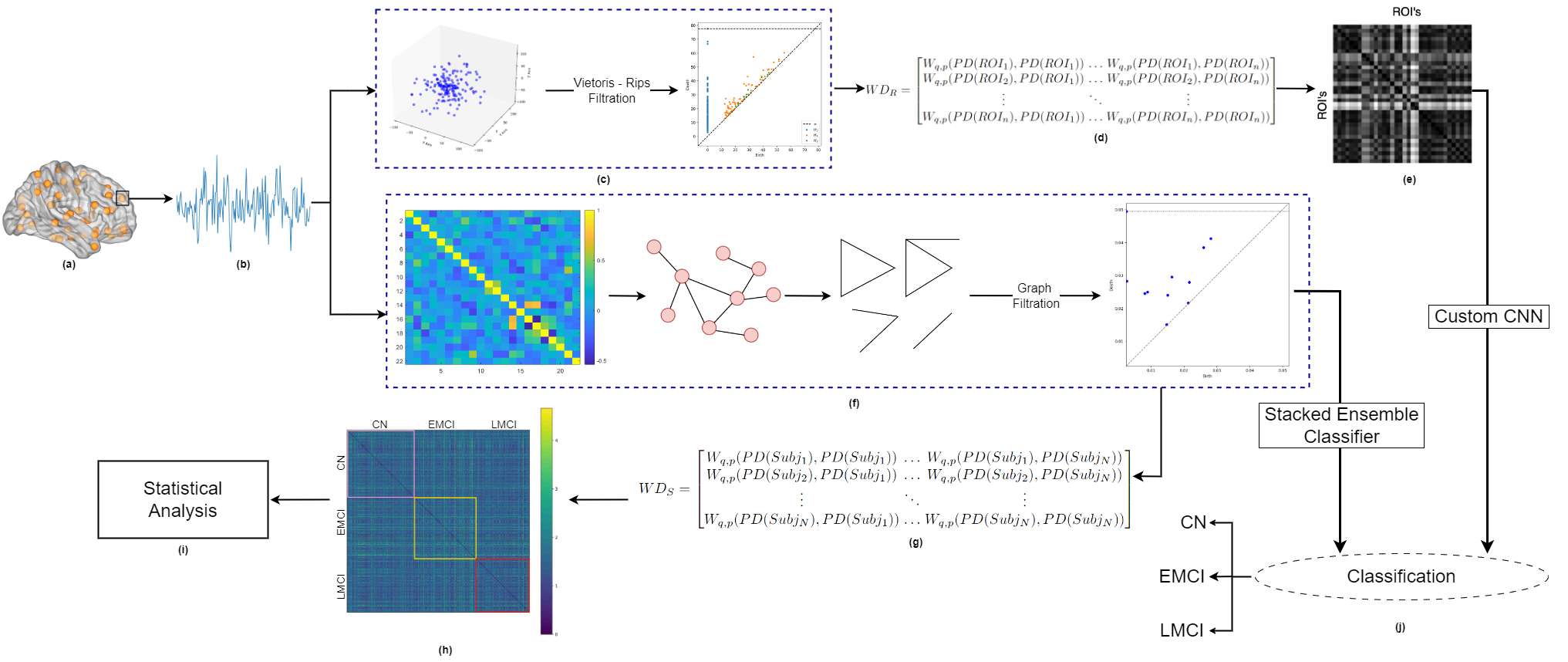}
    \caption{
Block diagram showing the Proposed Methodology.
}
    \label{fig:Methodology}
\end{figure}
The block diagram illustrating the proposed methodology is shown in Figure~\ref{fig:Methodology}. This study compares the classification performance achieved using two different filtration choices for constructing persistence diagrams: (i) Vietoris-Rips filtration and (ii) graph filtration. The analysis begins by extracting time series data from resting-state fMRI volumes. To generate persistence diagrams using Vietoris-Rips filtration, the 1D fMRI time series is first transformed into a 3D point cloud. In contrast, for graph filtration, correlation analysis is performed directly on the extracted 1D fMRI time series data, using both marginal and partial correlations to construct a positively correlated graph from distinct brain regions (nodes). The adjacency matrix generated from this graph is then used to compute persistence diagrams via graph filtration. 
To quantify topological changes, we employ the Wasserstein distance metric. These changes are analyzed in two distinct ways: (i) across subjects for a specific region of interest (ROI) when using persistence diagrams generated from graph filtration, and (ii) across ROIs for a given subject when using persistence diagrams derived from Vietoris-Rips filtration. 
Finally, the classification of healthy individuals, EMCI, and LMCI is carried out using two feature sets: (a) the top ten most persistent homology features obtained from graph filtration, and (b) inter-ROI Wasserstein distance features derived from the Vietoris-Rips filtration.
Each step of the proposed methodology is described in detail in the following sub-sections.

\subsection{Extraction of fMRI time series}

\begin{figure}[h]
    \centering
    \fbox{\includegraphics[width=0.6\textwidth]{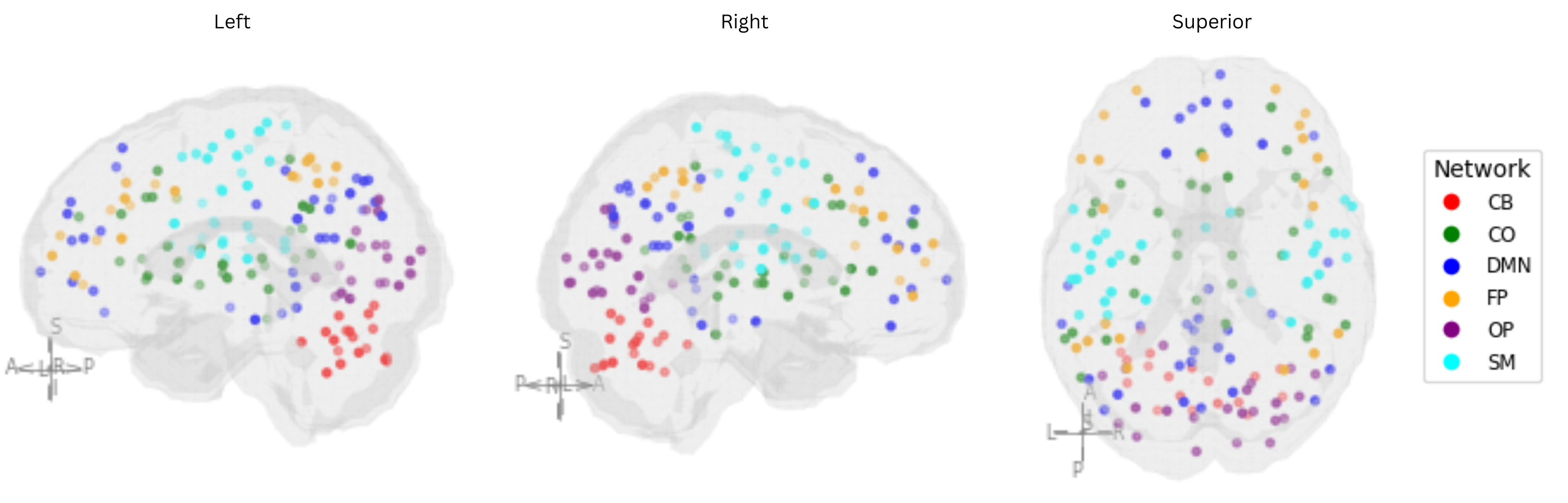}}
    \caption{
Image showing Dosenbach’s ROIs for the six networks. 
}
    \label{fig:Network Plot}
\end{figure}

The fMRI time series provides insights into the temporal dynamics of brain activity, allowing for an in-depth analysis of how various brain regions interact over time. In this study, specific brain regions are carefully selected from Dosenbach’s ROIs to extract fMRI time series, aiming to identify significant patterns and differences between MCI sub-types and healthy controls. The Dosenbach’s ROIs (\cite{dosenbach_2010_prediction}), consisting of 160 regions, are categorized into six distinct brain networks: cerebellum (CB) comprising 18 nodes, cingulo-opercular (CO) with 32 nodes, default mode network (DMN) consisting of 34 nodes, fronto-parietal (FP) comprising 21 nodes, occipital (OP) with 22 nodes, and sensorimotor (SM) consisting of 33 nodes. Figure~\ref{fig:Network Plot} displays the different ROIs for these six distinct networks. These networks encompass various interconnected brain regions, each associated with particular cognitive, sensory, and motor functions. Given that these functions may exhibit unique disruption patterns at different stages of MCI, the analysis of all six networks offers a comprehensive assessment of network-specific changes. These changes could potentially serve as distinct biomarkers for different stages or types of cognitive impairment. A representative 5mm radius sphere centered at each voxel location was used to generate the time series ${v_{t}, t = 1, 2, . . . , N}$. 
The fMRI preprocessing was carried out using FSL's FEAT. Briefly, the steps involved discarding the initial 10 volumes, applying MCFLIRT for motion correction, performing brain extraction, applying a 5mm spatial smoothing kernel, and conducting high-pass temporal filtering. Additionally, the fMRI data were registered (using a 12-degree-of-freedom linear transformation) to the corresponding structural image and then to MNI152 space. As per standard practices, the mean contributions of motion correction parameters, as well as CSF and WM signals—treated as nuisance variables—were regressed out during preprocessing.

\subsection{Persistent homology using Vietoris-Rips filtration}

Efficiently creating a point cloud representation from 1D time series data is a crucial step in computing persistent homology (\cite{sw1,sw2}), which is essential for analyzing the intrinsic topological properties of MCI. To achieve this, the study uses sliding window embedding (SW) with an embedding dimension \( M = 2 \) and a time lag \( \tau = 1 \), converting the fMRI time series \( v_t \) into 3D point clouds \( S \). The sliding window length is set to 3, minimizing noise and enhancing interpretability.
The sliding window embedding of a function $f$ based at $t \in \mathbb{R}$ into 
$\mathbb{R}^{M+1}$ is represented as follows (Equation~\ref{eq:sliding}):
\begin{equation}
 \mathcal{SW}_{M,\tau} f: \mathbb{R} \rightarrow \mathbb{R}^{(M+1)}, \quad t \rightarrow 
\begin{bmatrix}
 f(t) \\
 f(t+\tau) \\
 f(t+2\tau) \\
 \end{bmatrix}
 \label{eq:sliding}
\end{equation}
By selecting different values of \( t \), a sliding window point cloud is generated, representing the function \( f \) in 3D space. This approach is supported by literature indicating the effectiveness of sliding window methods in capturing dynamic functional connectivity in rs-fMRI.
In this study, for embedding dimension $M = 2$, the point cloud of the fMRI time series is represented by Equation~\ref{eq:pointcloud}:
\begin{equation}
 \mathcal{S} = \{v_{i} : i = 1, \ldots, N, \ v_i \in \mathbb{R}^3\}
 \label{eq:pointcloud}
\end{equation}
Each point \( v_i \) in the point cloud \( S \) is mapped to a vector in \( \mathbb{R}^3 \), maintaining consistency with the spatial dimensions of the fMRI data.

Persistent homology is then computed from these 3D point clouds to extract topological features. This is done by constructing a series of simplicial complexes using Vietoris-Rips filtration and calculating their homological features. These features capture the underlying topological structure in the data, highlighting meaningful patterns and differences between MCI and its subtypes. We exploit the information encoded in \textit{persistence diagram} to analyze the differences in topology of brain networks of individuals with MCI from HC.
Persistence diagram encodes the persistence features in data across the filtration parameter range as a collection of points in the two-dimensional Euclidean space $\mathbb{R}^2$. A common approach for constructing a filtration from a point cloud is through the Vietoris–Rips complex. This complex is generated from the point cloud by connecting any subset of points whose pairwise distances fall within a specified threshold, creating a simplex.
Thus, filtration is a collection $\mathcal{F}= \{F_{\epsilon}\}_{\epsilon \geq 0} $ of spaces with $F_{\epsilon} \subset F_{\epsilon^{\prime}}$
continuous $\forall \epsilon \leq \epsilon^{\prime}$. The $i^{th}$ persistence diagram of $\mathcal{F}$ is a multiset $dgm_{i}(\mathcal{F}) \subset \{(p,q) \in [0, \infty] \times [0, \infty] \mid 0 \leq p < q \} $ where each pair $(a,b) \in dgm_{i}(\mathcal{F})$ encodes a \textit{i}-dimensional topological feature, in other words Betti descriptors\footnote{In algebraic topology, the topological features of a space are represented as \emph{holes} or \emph{cycles} in various dimensions. The number of $k$-dimensional holes in a $d$-dimensional simplicial complex (with $k \leq d$) is denoted by the Betti number $\beta_k$ or $H_k$. Thus, $0$-dimensional holes $(\beta_0 or H_0)$ correspond to connected components, $1$-dimensional holes $(\beta_1 or H_1)$ represent tunnels (or loops) and $2$-dimensional holes $(\beta_2 or H_2)$ are voids.} associated with a simplicial complex that born at $F_{b}$ and dies at $F_{d}$. Here, persistent homology features were computed for 0-dimension, 1-dimension, and 2-dimension separately. The quantity $(d-b)$ is the persistence of the feature, and typically measures significance across the filtration. In our study, given a time series ($V_t$) the sliding window point cloud $\mathcal{SW}_{M,\tau} f$ is computed which is in a metric space $(X, M_{X})$. The Rips filtration $\mathcal{VR}(X,M_{X})$ is derived from the Vietoris–Rips complex $VR_{\epsilon}(X,M_{X})$, computed at each scale $\epsilon \geq 0$. The mathematical expression for computing Rips filtration is depicted in Equation~\ref{eq:vr}
\begin{multline}
 \mathcal{VR}(X,M_{X}):= \{VR_{\epsilon}(X,M_{X})\}_{\epsilon \geq 0}, where \\
 VR_{\epsilon}(X,M_{X}) := \{\{x_{0},..., x_{n}\} \in X \mid \max\limits_{0\leq i, j \leq n} 
M_{X}(x_{i},x_{j}) < \epsilon, \textit{n} \in \mathbb{N}
\label{eq:vr}
\end{multline}
The birth-death pairs $(b,d)$ in the Rips persistence diagrams $dgm_{i}^{\mathcal{VR}} (X):= dgm_{i}\mathcal{VR}(X, M_{X})$ reveal the underlying topology of space $X$. The points $(b,d)$ in $dgm_{i}^{\mathcal{VR}} (X)$ with large persistence values $(d-b)$ suggest the most persistent topological features of the continuous space where $X$ is concentrated.

\subsection{Persistent homology using graph filtration}

\subsubsection{Graph construction using marginal and partial correlation}

A network consists of nodes (vertices) and links (edges) that connect pairs of nodes. In the context of brain networks, the nodes correspond to distinct brain regions, while the edges represent the strength of connectivity between them. Mathematically, this network can be represented as an undirected graph \(G = (V, E)\), where \(V\) is the set of nodes and \(E\) is the set of edges. Each edge \(l_{ij} \in E\), connecting node \(i\) with node \(j\), has an associated weight—positive or negative—that reflects the temporal correlations in brain activity between the two regions. The entire network is captured in a symmetric adjacency matrix of size \(N \times N\), where \(N\) is the number of nodes. In this matrix, each entry \((i, j)\) indicates the strength of the edge between nodes \(i\) and \(j\). In our study, these edge weights are computed using both marginal and partial correlations.

Correlation analysis has long been a prevalent method for investigating connectivity between brain regions \cite{kim_2015_testing, wang_2016_an}. Most studies have relied on Pearson correlation, also known as marginal correlation, which captures only the marginal associations between network nodes. However, using Pearson correlation alone is insufficient for brain connectivity analysis, as it does not account for the true or direct connections between nodes. For example, significant correlations between two nodes, X and Y, may arise due to their shared connection with a third node, Z, even when X and Y are not directly connected \cite{kim_2015_testing}. This reliance on marginal correlation complicates the distinction between network edges representing true connectivity and those influenced by confounding factors.
To overcome this limitation, partial correlation has emerged as a powerful statistical technique \cite{smith_2012_the, kim_2015_testing, wang_2016_an}. Partial correlation estimates the relationships between nodes while controlling for the spurious effects of all other nodes in the network, providing a more accurate measure of direct connectivity. A zero value in partial correlation indicates an absence of direct connectivity between node pairs. Extensive literature demonstrates that partial correlation is one of the most effective techniques for identifying true functional connectivity between network nodes, often outperforming traditional methods and showing high sensitivity in revealing genuine network connections \cite{kim_2015_testing, erb_2020_partial, zhang_2021_partial}. 
In this study, we focus exclusively on positively correlated networks for further analysis. Positive networks are constructed when both marginal and partial correlation edge strengths indicate a positive association. The visualization of a positively correlated network for one subject is presented in Figure~\ref{fig:Methodology}(f).

\subsubsection{Graph Filtration}

Consider a weighted graph $G = (V, E)$, where $V$ and $E$ represent the sets of vertices and edges, respectively. The weight of an edge $e$ is denoted by $w(e)$. A filter function $f : G \to \mathbb{R}$, on $G$, is defined as follows:
\begin{itemize}
    \item for an edge $e \in E$, the value of $f(e)$ is set to $w(e)$,
    \item for a vertex $v \in V$, the value of $f(v)$ is obtained by selecting the minimum among all edge weights associated with the edges incident on $v$.
\end{itemize}

For each $r \in \mathbb{R}$, the subgraph $G_{\leq r}$ of $G$ is defined as the collection of vertices and edges in $G$ with $f$-values at most $r$. Similarly, the subgraph $G_{\geq r}$ consists of vertices and edges with $f$-values greater than or equal to $r$. Let $w_1 \leq w_2 \leq \dots \leq w_{|E|}$ be the weights of the edges in $G$, in non-decreasing order, where $|E|$ denotes the number of edges in $G$. This arrangement yields two sequences of graphs, each referred to as a filtration. The sequence $\emptyset = G_{\leq 0} \subseteq G_{\leq w_1} \subseteq G_{\leq w_2} \subseteq \dots \subseteq G_{\leq w_n} = G$ is called the sublevel set filtration of $f$, and the sequence $G_{\geq w_n} \subseteq G_{\geq w_{n-1}} \subseteq \dots \subseteq G_{\geq w_0} = G$ is referred to as the superlevel set filtration of $f$.
The filtrations provide a basis for computing the persistent topological features that exist within the graph. 

\begin{figure}[h]
    \centering
    \includegraphics[width=\textwidth]{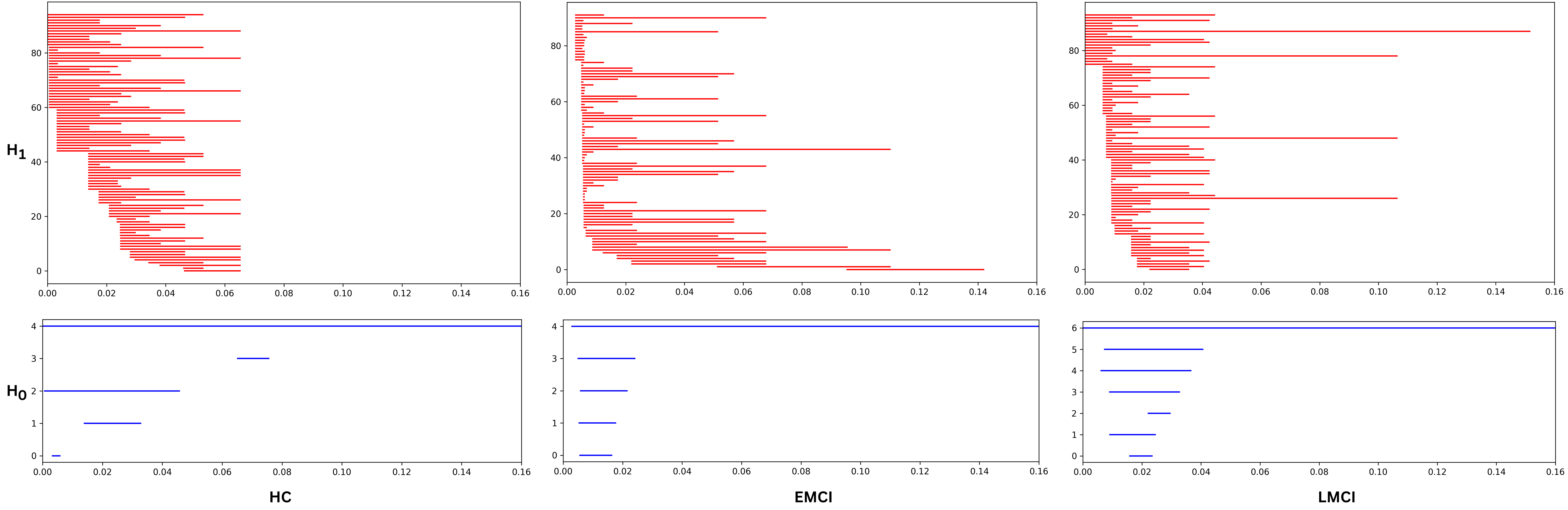}
    \caption{
The persistence barcode for a representative subject for Healthy, EMCI and LMCI group for network OP.
}
    \label{fig:Persistence Barcodes}
\end{figure}

In this analysis, we generate persistence diagrams for all subjects across six distinct brain networks to effectively capture and compare the topological features inherent to each group. This approach provides valuable insights into the structural differences among the groups. 
A visual representation of these raw topological features is illustrated through persistence barcodes, as shown in Figure~\ref{fig:Persistence Barcodes}. This figure presents the persistent homology features for one representative subject from each of the three groups. Each horizontal bar in the barcodes corresponds to a specific topological feature, with the start and end points of each bar indicating the birth and death of that feature, respectively. The length of the bars reflects the persistence of the features; longer bars are interpreted as being more significant, suggesting that these features are more robust and likely to contribute meaningfully to the underlying topology of the network. This visualization allows for an intuitive understanding of the persistence of various topological features and their relevance in distinguishing between the brain networks of the different subject groups. 

\subsection{Quantification of persistence diagrams using Wasserstein distance}

To measure the dissimilarity between persistence diagrams, we utilize the Wasserstein distance. This metric is particularly effective in the context of persistence diagrams because it captures not only the differences in the locations of points but also their distribution within the metric space, offering a more comprehensive understanding of the represented topological features \cite{berwald2018computingwassersteindistancepersistence}. 
The Wasserstein distance has several advantages over alternative metrics, such as the bottleneck distance. One limitation of the bottleneck distance is its insensitivity to the details of the bijection beyond the most distant pair of corresponding points, which can result in a loss of important information. For this reason, our study focuses on the Wasserstein distance for quantification. By quantifying the differences in topological features between subjects, the Wasserstein distance provides a robust foundation for distinguishing between various subtypes of MCI and healthy individuals. This approach enhances our ability to compare topological changes in brain activity effectively.

\subsubsection{\textit{Network-Specific} Inter-Subject Wasserstein distance}
\label{sec:InterSubject}

Network-specific inter-subject Wasserstein distance was computed for the two Betti descriptors: $H_0$ and $H_1$, generated using graph filtration. The Wasserstein distance between the persistence diagrams of the positively correlated connectivity graphs corresponding to subjects \({S_{1}}\) and \({S_{2}}\), quantifies the structural similarity in the topology of their brain networks. A low Wasserstein distance indicates that the subjects ${S_{1}}$ and ${S_{2}}$ have similar topological structures in their brain networks, suggesting functional synchrony or comparable activity between their brain regions. Conversely, a high Wasserstein distance signifies greater topological differences between the subjects' brain networks, implying distinct connectivity patterns or functional dissociation. This difference may point to abnormalities in brain connectivity, potentially associated with neurological or psychiatric conditions.
In the context of the three groups (HC, EMCI, and LMCI), analyzing these distances across brain networks such as CB, CO, DMN, FP, OP, and SM helps to reveal MCI specific differences in brain network organization. 
To analyze differences in topological patterns across all subjects from the three groups (HC, EMCI, and LMCI) for each of the two Betti descriptors ($H_0$ and $H_1$), a network-specific analysis is performed for the six brain networks (CB, CO, DMN, FP, OP, SM). This is illustrated by the Wasserstein distance matrix ($WD_S$), which has dimensions N × N, where N represents the number of subjects. Equation~\ref{eq:WD_S} provides the mathematical representation of the $WD_S$ matrix. In this matrix, each element $WD_S$(i, j) reflects the Wasserstein distance between the persistence diagrams of subject i and subject j for a particular network.

\begin{equation}
    WD_S=\begin{bmatrix}
    W_{q,p}(PD(Subj_1), PD(Subj_1)) & \dots & W_{q,p}(PD(Subj_1), PD(Subj_N)) \\
    W_{q,p}(PD(Subj_2), PD(Subj_1)) & \dots & W_{q,p}(PD(Subj_2), PD(Subj_N)) \\
    \vdots & \ddots & \vdots \\
    W_{q,p}(PD(Subj_N), PD(Subj_1)) & \dots & W_{q,p}(PD(Subj_N), PD(Subj_N)) \\
    \end{bmatrix}
    \label{eq:WD_S}
\end{equation}

A sample image as obtained from $WD_S$ for one specific network is shown in Figure~\ref{fig:Methodology}(h)

\subsubsection{\textit{Subject-Specific} Inter-ROI Wasserstein distance}
\label{sec:InterROI}

For each brain network, subject-specific inter-ROI Wasserstein distance was computed for each of the three Betti descriptors: $H_0$, $H_1$, and $H_2$, generated using Vietoris-Rips filtration. The Wasserstein distance between fMRI time series from two brain regions reflects the similarity in their neural activity patterns. A low Wasserstein distance suggests that the two fMRI time series share similar activity patterns over time, indicating that the regions are functionally synchronized and likely engaged in coordinated activity. Conversely, a high Wasserstein distance implies distinct neural activity patterns between the two regions, which may be functionally dissociated or independent. Additionally, high Wasserstein distances could signal abnormalities in functional connectivity between the regions, potentially pointing to neurological or psychiatric disorders.
The pairwise-ROI distance matrix ($WD_{ROI}$), with dimensions \( n \times n \), where \( n \) denotes the number of regions of interest (ROIs) in a specific brain network, captures the interaction between different ROIs. Each entry in the matrix \( PR(i, j) \) represents the Wasserstein distance between the persistence diagrams of ROI \( i \) and ROI \( j \). The mathematical representation of the matrix $WD_{ROI}$ is shown in 
Equation~\ref{eq:WD_{ROI}}.
Calculating both the persistent homology and the Wasserstein distance matrices (PR and PS) is computationally demanding. To address this challenge, we employed high-performance computing (HPC) resources, specifically an Intel(R) Xeon(R) Gold 6240 CPU @ 2.60GHz with dual CPUs and 192 GB of memory. 

\begin{equation}
 WD_{ROI}=\begin{bmatrix}
 W_{q,p}(PD(ROI_1), PD(ROI_1)) & \dots & W_{q,p}(PD(ROI_1), PD(ROI_n)) \\
 W_{q,p}(PD(ROI_2), PD(ROI_1)) & \dots & W_{q,p}(PD(ROI_2), PD(ROI_n)) \\
 \vdots & \ddots & \vdots \\
 W_{q,p}(PD(ROI_n), PD(ROI_1)) & \dots & W_{q,p}(PD(ROI_n), PD(ROI_n)) \\
 \end{bmatrix}
 \label{eq:WD_{ROI}}
\end{equation}
Figure~\ref{fig:Methodology}(e) illustrates the variability in Wasserstein distance among all ROI pairs for a single representative subject.

\subsection{Classification}

Classification is conducted using two distinct sets of features: \\
\textit{Set-1:} This set comprises raw features derived from the persistence diagrams, specifically focusing on the lifespan of topological descriptors calculated using graph filtration. These features provide insights into the persistence and significance of various topological structures within the brain networks. \\
\textit{Set-2:} This set includes the inter-region of interest (ROI) Wasserstein distances computed using Vietoris-Rips filtration. These distances quantify the dissimilarities between the persistence diagrams of different brain regions, capturing essential topological information about connectivity patterns.

For classification, we employ a stacked ensemble classifier for the first set of features, leveraging its ability to combine multiple learning algorithms to improve predictive performance. In contrast, we utilize a custom convolutional neural network (CNN) model for the second set of features, which is designed to effectively capture spatial hierarchies and complex patterns within the Wasserstein distances. This dual approach allows us to maximize the strengths of both feature sets, enhancing our overall classification accuracy.

\subsubsection{Stacked Ensemble Classifier}

An ensemble classifier integrates the predictions from multiple models to achieve a more accurate and robust classification than any individual model could provide. By leveraging the strengths of various algorithms, this approach reduces the likelihood of errors and enhances the model's ability to generalize to new data. In our case, the ensemble method was employed to capitalize on the complementary advantages offered by different classifiers.
The classification process is based on the top ten most persistent homology features, where the lifespan of each feature in $H_{0}$ and $H_{1}$ serves as the primary input for training. The dataset is divided into training (80\%) and testing (20\%) subsets, ensuring that the model's performance could be assessed on unseen data after training, thereby providing a realistic evaluation of its generalization capabilities.
First, feature selection is performed using Recursive Feature Elimination (RFE), a technique that systematically removes the least important features to identify the most relevant ones. A Random Forest classifier was utilized within RFE to rank and eliminate features, ultimately pinpointing the five most significant features from the original ten persistent homology points.
Once the key features are selected, a stacking ensemble was constructed, incorporating a diverse array of base classifiers, including Support Vector Classifier, Random Forest, Gradient Boosting, XGBoost, AdaBoost, Extra Trees, Logistic Regression, K-Nearest Neighbors, LightGBM, and CatBoost. Each of these models was trained independently on the training dataset, allowing them to capture different patterns and relationships within the data that a single model might overlook.
To ensure a robust assessment of model performance, all base models were evaluated using stratified K-Fold cross-validation $(K=5)$. This technique guarantees that each fold maintains the same proportion of samples from each class, effectively addressing potential issues with imbalanced data and providing a more reliable estimate of model accuracy.
Following this, the classifiers were ranked based on their cross-validation accuracy scores, and the top-performing models were selected for inclusion in the final stacking ensemble. The stacking classifier was assembled using the five highest-ranked models, with a Random forest classifier designated as the meta-model to aggregate the predictions from the base classifiers. The predictions from these five top models on the training dataset were then used as inputs for the meta-classifier, allowing the stacking approach to leverage the unique strengths of each individual classifier, thereby further enhancing overall performance.
The model was trained across four scenarios: (1) HC Vs. EMCI; (2) HC Vs. LMCI; (3) EMCI Vs. LMCI for the ADNI dataset; and (4) HC Vs. MCI for the in-house TLSA dataset. 
Finally, the performance of the model is tested on unseen data, and evaluation metrics such as accuracy, confusion matrix, and classification report were generated for the test set. This comprehensive approach ensures that our classification model is both robust and effective in differentiating between various cognitive states.

\subsubsection{Proposed CNN framework}

\begin{figure} [h]
\centering
\includegraphics[width=0.7\textwidth]{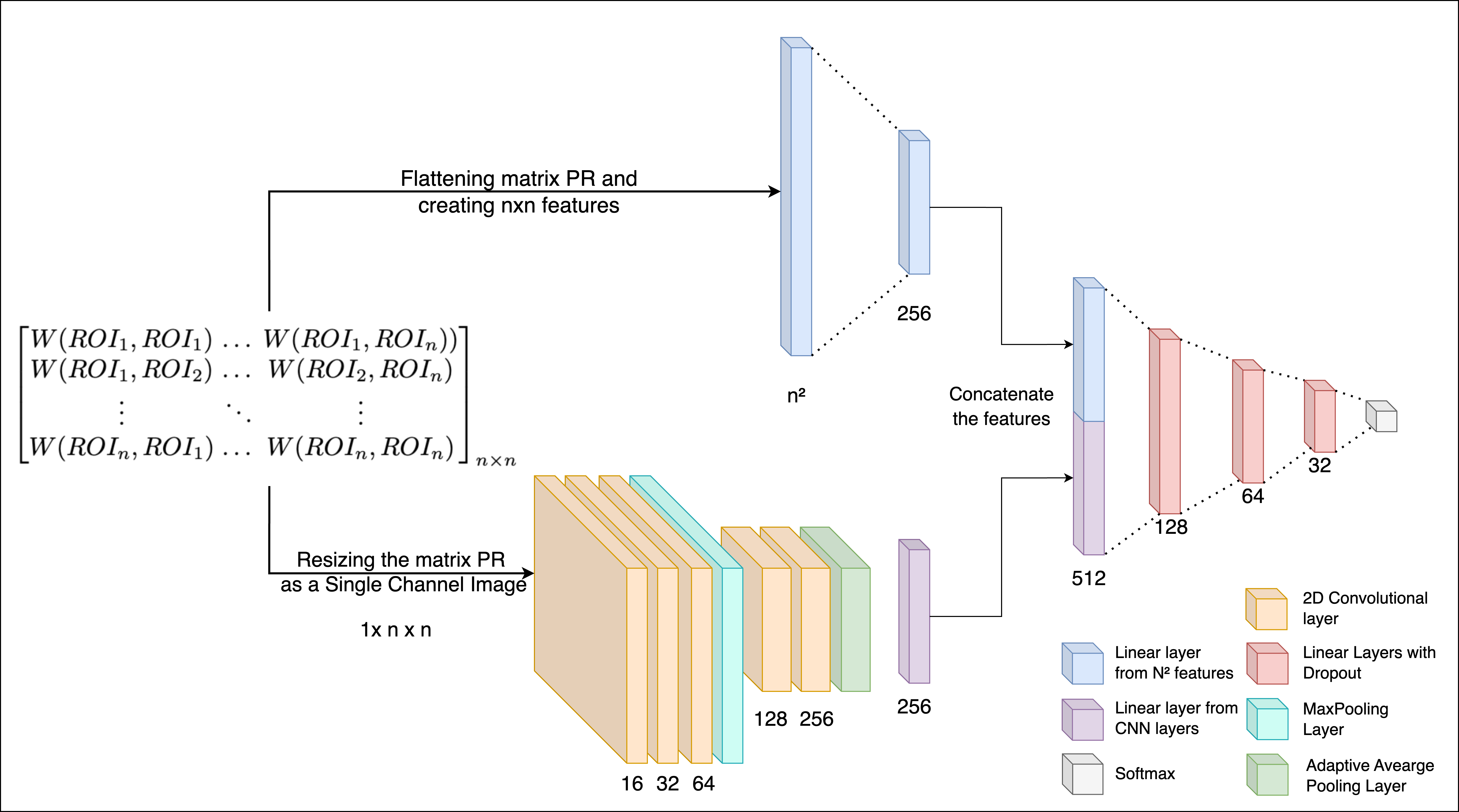}
\caption{The deep learning architecture for classification using persistent diagrams generated with Vietoris-Rips filtration}
\label{fig:ICPRmodel}
\end{figure}

The study incorporates both 1D and 2D features derived from Wasserstein distances, which capture inter-ROI interactions for each subject, into a classification framework utilizing a conventional CNN. For classification, a subject-specific $WD_{ROI}$ matrix ($n \times n$) is employed. The proposed CNN architecture, illustrated in Figure \ref{fig:ICPRmodel}, combines 1D features extracted from each ROI pair in the $WD_{ROI}$  matrix with 2D CNN-derived features.
The proposed classification model operates in two steps. First, the Wasserstein distance matrix is flattened to generate 1D features, focusing on pairwise relationships. Then, 2D features are extracted from the matrix using CNN layers, capturing local patterns and spatial hierarchies (Figure~\ref{fig:Methodology}e). These features are concatenated to form a unified feature vector, integrating information from both linear and convolutional layers. This unified vector is processed through several dense layers with dropout for regularization, minimizing the risk of overfitting. The combination of 1D and 2D features results in a richer, more diverse feature space that captures various aspects of the data—1D features represent sequential or linear relationships, while 2D features capture spatial or topological relationships. This integration offers multiple benefits, including a comprehensive feature space for classification, improved learning from different perspectives, noise and artifact mitigation, and the ability to capture both local and global patterns. For the 2D features, the model includes three CNN layers with 16, 32, and 64 filters, followed by a max-pooling layer and an additional convolutional block with two CNN layers containing 128 and 256 filters. All CNN layers use a kernel size of 3. A global average pooling layer then condenses each feature map into a single value, forming a linear feature vector, with ReLU activation functions applied throughout. Simultaneously, the 1D features of the $WD_{ROI}$  matrix ($n^2 \times 1$) are processed through a linear layer, reducing them to 256 features. The 256-dimensional feature vectors from both the linear layer and the 2D CNN are concatenated and passed through a series of fully connected layers with sizes 128, 64, and 32, each incorporating a dropout rate of 0.2. The final layer employs a softmax activation function for classification.

Each Betti descriptor ($H_{0}$, $H_{1}$, and $H_{2}$) from every brain network is independently analyzed. The model is trained over 100 epochs using the Adam optimizer with a learning rate of 0.001, with an 80-20 train-test split on unique subjects. Cross-Entropy loss is used for training. The classification tasks include (i) MCI versus HC (for both ADNI and our in-house dataset), (ii) EMCI versus HC (ADNI), (iii) LMCI versus HC (ADNI), and (iv) EMCI versus LMCI (ADNI). The entire experiment is conducted on $24$ GB NVIDIA A5000 GPUs and the PyTorch deep learning framework.

\section{Results}

This study utilizes persistent homology to investigate variations in brain network topology between healthy individuals and those diagnosed with MCI having different stages (Early/Late). Positively correlated graphs are generated for six classical brain networks— CB, CO, DMN, FP, OP, and SM—derived from 160 Dosenbach ROIs, shown in Figure~\ref{fig:Network Plot}. These brain connectivity graphs are constructed based on the rs-fMRI time series of each network, considering only instances where both marginal and partial correlations exhibit positive values. 

First, graph filtration is employed to compute persistent homology for dimension-0 and -1 across each of the six brain network. The resulting persistence diagrams are then examined to identify key topological features. As outlined in the methodology, each point in a persistence diagram represents a specific topological feature, where the difference between the ``birth" and ``death" values signifies the lifespan or persistence of a feature. Longer persistence indicates more prominent or stable topological characteristics. 
Figure ~\ref{fig:Persistence Barcodes} visually illustrates significant differences in topological features among the three groups: HC, EMCI, and LMCI. The persistence barcodes for one representative subject from each group are shown for both dimension-0 (connected components) and dimension-1 (loops). 
The control group exhibits fewer topological features, and these features tend to have shorter lifespans, indicating more transient or less complex network structures. Additionally, there is minimal variation in the persistence of these features, suggesting more uniform brain network topology across healthy individuals.
In contrast, the EMCI group displays a wider range of topological features. While some features have short lifespans, similar to those in the HC group, EMCI also shows several features that persist for longer duration. This suggests that EMCI individuals have more complex and diverse topological structures than HC, potentially reflecting early-stage disruptions in brain network connectivity.
The LMCI group, on the other hand, presents a distinctive pattern. Most features in the persistence diagrams have relatively short lifespans, indicating that the majority of topological structures in their brain networks are transient or unstable. However, LMCI also demonstrates a few long-persisting features, which last longer than those observed in both the HC and EMCI groups. These features may represent more pronounced or severe alterations in brain connectivity, characteristic of late-stage MCI.

\subsection{Significant topological differences are obtained in raw persistence features between HC and MCI-subtypes}
As discussed in Section~\ref{sec:InterSubject}, inter-subject Wasserstein distances (shown in Figure~\ref{fig:Methodology}(h)) are computed from the raw persistence features derived through graph filtration to assess dissimilarities between the persistence diagrams of healthy and diseased groups.
\setlength{\tabcolsep}{4pt}
\begin{table} [h]
\begin{center}
\caption{Statistical significance of Wasserstein distance, computed using top 10 most persistent features, between Healthy Control (HC), early MCI (EMCI), and late MCI (LMCI) for different brain networks in ADNI data and TLSA data.}
\vspace{10pt}
\label{table:WD ADNI}
\renewcommand{\arraystretch}{1.5}
\begin{tabular}{|c|c|c|c|c|c|}
    \hline
     \textbf{Network}  &\textbf{Homology} &\multicolumn{3}{c|}{\textbf{ADNI Data}} & \textbf{TLSA Data}\\ \cline{3-6}
     & \textbf{Dimension} & \textbf{HC Vs. EMCI} & \textbf{HC Vs. LMCI} & \textbf{EMCI Vs. LMCI} & \textbf{HC Vs. MCI}\\ 
    \hline
    \multirow{2}{*}{} CB & $H_0$ & $p<0.001$ & $p<0.001$ & $p<0.001$  &$p<0.001$ \\ \cline{2-6}
     & $H_1$ & $p<0.001$  & $p<0.001$  & NS &$p<0.001$  \\ \hline
    \multirow{2}{*}{} CO & $H_0$ & $p<0.001$  & $p<0.001$  & $p<0.001$ & NS\\ \cline{2-6}
     & $H_1$ & $p=0.01$ & NS & NS &$p<0.001$ \\ \hline
    \multirow{2}{*}{} DMN & $H_0$ & $p<0.001$ & $p<0.001$ & $p<0.001 $ &$p<0.001$ \\ \cline{2-6}
     & $H_1$ & $p<0.001$  & $p<0.001 $ & $p<0.001 $  &$p=0.03$\\ \hline
   \multirow{2}{*}{} FP & $H_0$ & $p=0.02$ & $p<0.001 $  & $p<0.001 $ &$p<0.001$  \\ \cline{2-6}
     & $H_1$ & $p<0.001$  & $p<0.001$  & $p<0.001$  &NS\\ \hline
    \multirow{2}{*}{} OP & $H_0$ & $p<0.001 $  & $p=0.001$ & $p<0.001 $ &$p<0.001$ \\ \cline{2-6}
     & $H_1$ & $p=0.001$ & $p<0.001 $  & $p<0.001 $  &$p<0.001$  \\ \hline
    \multirow{2}{*}{} SM & $H_0$ & NS & $p<0.001 $  & $p<0.001 $ &$p=0.008$ \\ \cline{2-6}
     & $H_1$ & $p<0.001 $ & NS & $p<0.001$ &NS\\ \hline
\end{tabular}
\end{center}
\end{table}

 In this study, the Wasserstein distance serves as a metric for comparing the raw persistent homology features of brain connectivity graphs, providing valuable insights into structural variations across different subjects. In order to determine whether the topological features as obtained through persistent homology differ significantly between study groups, we tested the hypothesis that the difference in mean distribution of Wasserstein distances is statistically significant between healthy individuals and MCI groups. For this, the Wilcoxon rank-sum test is conducted at a 95\% confidence interval for each of the six brain networks. Table~\ref{table:WD ADNI} presents the statistical results for the inter-subject Wasserstein distance computed from the 0- and 1-dimensional raw persistent homology features for both the ADNI and TLSA datasets. The results show statistically significant differences $(p<0.001)$ in most persistence features across all six networks between healthy and MCI groups, as well as between different MCI sub-types.

 \subsection{Classification using top ten most persistent features}

The analysis of persistence diagrams reveals notable differences in the topological characteristics of MCI and its sub-types compared to HC, as derived through graph filtration. The differences in the number, persistence, and variation of topological features provide valuable insights into the progression of brain network changes from healthy to early and late MCI. 
As discussed earlier, features that persist longer tend to better capture the underlying topological structure of the graph compared to the features that do not persist longer and might actually be a result of noise. Therefore, for classification purpose, the top ten most persistent features of $H_{0}$ and $H_{1}$ are used separately to differentiate the three groups using stacked ensemble classifier. However, while comparing the classification performance between the two homology dimensions, the best classification accuracy is found with most persistent $H_{1}$ features, with 63.9\% in classifying EMCI vs. LMCI in the DMN. For our in-house TLSA data, the model classified HC Vs. MCI with the accuracy of 71.4\% in the DMN. The performance of the stacked classifier in classifying MCI sub-types and healthy controls is summarized in Table~\ref{table:Stack Classify}. While graph filtration successfully captures some degree of network topology, the model's limited performance suggests that the raw persistent homology features may not fully exploit the complexity of the underlying brain network's structure, particularly in more challenging classification tasks.
\setlength{\tabcolsep}{3pt}
\begin{table}[h!]
\begin{center}
\caption{The classification accuracy (in \%) as obtained across six functional brain networks using stacked ensemble classifier with top ten most persistent raw homology features for dimension-1}
\vspace{10pt}
\label{table:Stack Classify}
\renewcommand{\arraystretch}{1.5}
\begin{tabular}{|c|c|c|c|c|}
    \hline
     \textbf{Network}  &\multicolumn{3}{c|}{\textbf{ADNI Data}} & \textbf{TLSA Data}\\ \cline{2-5}
      & \textbf{HC Vs. EMCI} & \textbf{HC Vs. LMCI} & \textbf{EMCI Vs. LMCI} & \textbf{HC Vs. MCI}\\ 
    \hline
    \multirow{2}{*}{} DMN  & 53.6 & 50.0 & \textbf{63.9} & \textbf{71.4} \\ \hline
    \multirow{2}{*}{} FP  & 55.1 & 51.6   & 50.8 & 60.0  \\ \hline
    \multirow{2}{*}{} OP  & 57.9  & \textbf{56.0} & 49.0 & 50.0 \\ \hline
    \multirow{2}{*}{} SM  & 53.6 & 54.7  & 57.0 & 66.67 \\ \hline
    \multirow{2}{*}{} CO  & \textbf{61.8} & 53.0 & 54.0 & 60.0 \\ \hline
    \multirow{2}{*}{} CB  & 58.8 & 50.0 & 59.0  & 60.9 \\
    \hline
\end{tabular}
\end{center}
\end{table}

\subsection{Classification Using Inter-ROI Wasserstein Distance as Features}

In addition to graph filtration, Vietoris-Rips filtration is utilized to derive persistence diagrams. As previously discussed, Vietoris-Rips filtration applies persistent homology to dimensions 0, 1, and 2 on 3D point clouds constructed from rs-fMRI time series data. This process helps identify persistent topological features within the point clouds. To quantify the dissimilarities between persistence diagrams, the Wasserstein distance metric is employed. The inter-ROI Wasserstein distance ($WD_{ROI}$) is calculated for each homology dimension ($H_0$, $H_1$, and $H_2$) based on the persistence diagrams, and these distances are then used as input features in a convolutional neural network (CNN) model for classification.
The performance of the proposed CNN model, using inter-ROI Wasserstein distances as features, in classifying MCI and its subtypes is summarized in Table~\ref{table:CNN classify}. The CNN model achieves a classification accuracy of 85.7\% for the in-house TLSA MCI cohort. However, when classifying MCI subtypes from healthy controls, the accuracy drops to 70.8\% in distinguishing EMCI from HC and classification of LMCI from HC achieves an accuracy of 81.0\% , which is on par with the TLSA HC versus MCI classification. When distinguishing between EMCI and LMCI, the model attains an accuracy of 77.3\%.

\setlength{\tabcolsep}{4pt}
\begin{table}[h]
\begin{center}
\caption{The classification accuracy (in \%) as obtained across six distinct brain networks using CNN with Inter-ROI Wasserstein Distance as features. Split based on Unique subjects}
\vspace{10pt}
\label{table:CNN classify}
\renewcommand{\arraystretch}{1.5}
\begin{tabular}{|l|p{2cm}|p{2cm}|p{1cm}|p{1cm}|p{1cm}|p{1cm}|p{1cm}|p{1cm}|}
\hline
\textbf{Comparison} & \textbf{Dataset} & \textbf{Homology Dimension}& \textbf{DMN} & \textbf{FP} & \textbf{OP} 
& \textbf{SM} & \textbf{CO} & \textbf{CB} \\
\hline
HC vs EMCI & ADNI & $H_{0}$ & 
\textbf{70.8} & 56.0 & 56.7 & 57.8 & 61.4 & 67.2 \\
\cline{3-9}
& & $H_{1}$ & 70.2 & 56.3 & 66.0 & 66.7 & 61.9 & 60.9 \\
\cline{3-9}
& & $H_{2}$ & 66.6 & 60.0 & 69.1 & 66.52 & 55.4 & 66.1 \\
\hline
HC vs LMCI & ADNI & $H_{0}$ & 66.7 & 71.0 & \textbf{81.0} & 60.0 & 66.1 & 63.2 \\
\cline{3-9}
& & $H_{1}$ & 67.6 & 58.7 & 75.0 & 67.2 & 68.2 & 64.3 \\
\cline{3-9}
& & $H_{2}$ & 74.6 & 59.7 & 65.6 & 63.6 & 71.4 & 50.0 \\
\hline
EMCI vs LMCI & ADNI & $H_{0}$ & 69.8 & 54.5 & 68.2 & 69.9 & 62.9 & 63.3 \\
\cline{3-9}
& & $H_{1}$ & 58.1 & 75.4 & 65.7 & 70.0 & \textbf{77.3} & 58.3 \\ 
\cline{3-9}
& & $H_{2}$ & 58.7 & 71.9 & 57.7 & 66.7 & 61.4 & 71.6 \\ 
\hline
HC vs MCI &In-house & $H_{0}$ & \textbf{85.7} & 78.6 & 64.7 & 78.6 & 71.4 & 70.6 \\
\cline{3-9}
&TLSA & $H_{1}$ & 64.7 & 64.3 & 71.4 & 71.4 & 82.4 & 78.6 \\
\cline{3-9}
& & $H_{2}$ & \textbf{85.7} & 64.7 & 71.4 & 78.6 & \textbf{85.7} & 64.3 \\ 
\hline
\end{tabular}
\end{center}
\end{table}

\subsection{Comparison between Graph filtration Vs. Vietoris-Rips filtration}
\begin{figure} [h]
\centering
\includegraphics[width=0.9\textwidth]{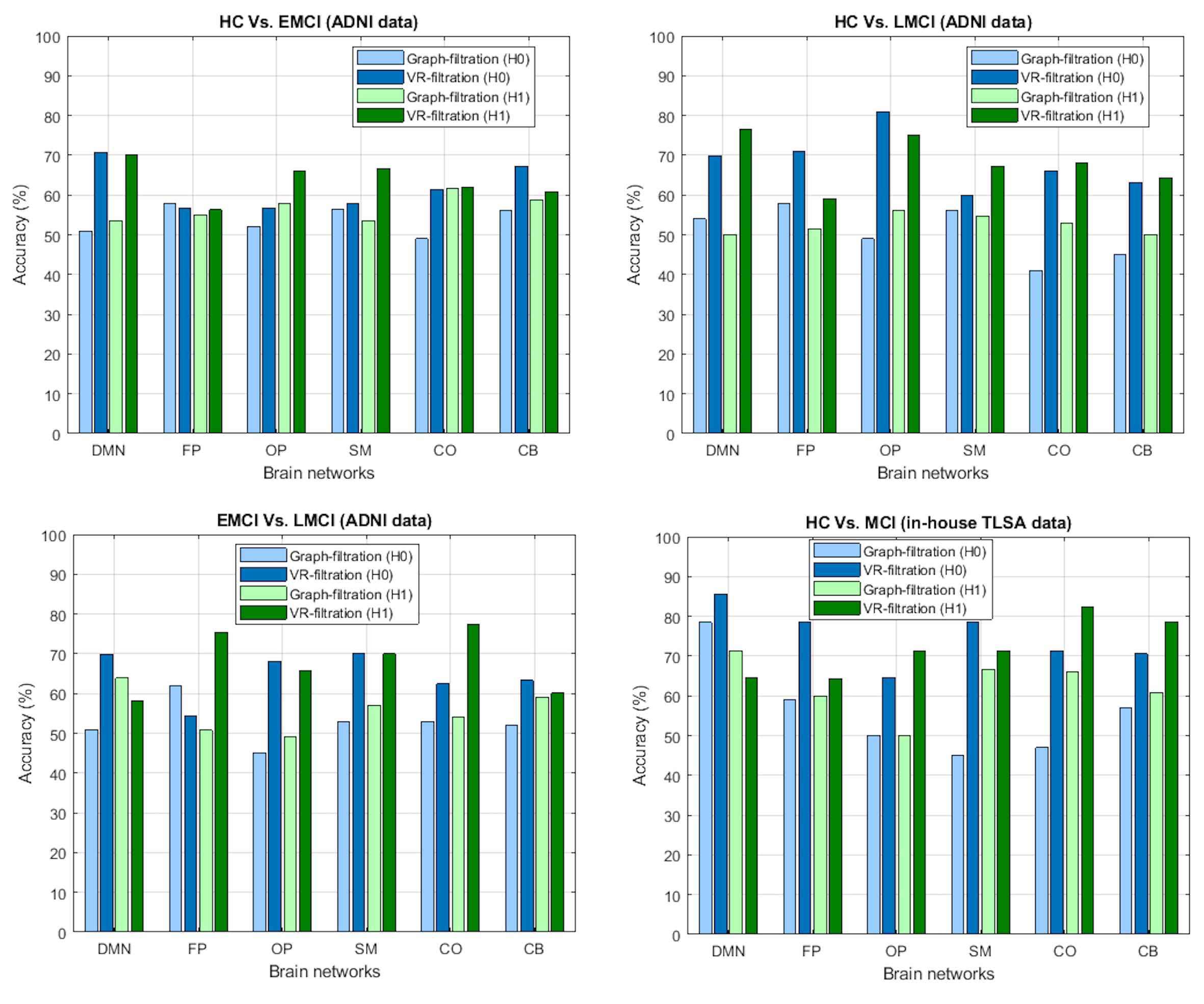}
\caption{Illustration of the comparative analysis between graph filtration and Vietoris-Rips filtration for both dimension-0 \((H_{0})\) and dimension-1 \((H_{1})\) in the classification of (A) healthy controls (HC) versus early mild cognitive impairment (EMCI), (B) HC versus late mild cognitive impairment (LMCI), (C) EMCI versus LMCI, and (D) HC versus mild cognitive impairment (MCI) based on TLSA data. The plot clearly highlights the superiority of Vietoris-Rips filtration over graph filtration, showcasing its enhanced ability to differentiate between these cognitive states. }
\label{fig:comp_plot}
\end{figure}

While comparing the classification performance between these two types of filtrations, it is seen that the proposed CNN model using Vietoris-Rips filtration and inter-ROI Wasserstein distance features outperformed the stacked ensemble classifier based on graph filtration in all classification tasks. The Vietoris-Rips filtration demonstrated better accuracy in differentiating between MCI subtypes and HC, indicating that it captures more detailed topological features of brain connectivity networks. This suggests that Vietoris-Rips filtration provides a more robust representation of brain network topology for MCI classification. The superior performance of Vietoris-Rips filtration when comparing these two filtration methods can be attributed to three key factors, which have both methodological and biological significance. This comparison is depicted in Figure~\ref{fig:comp_plot} through a bar diagram and is further elaborated in the subsequent subsections.

\subsubsection{Point cloud representation Vs. fixed graph structure}
Vietoris-Rips filtration operates on 3D point clouds generated from rs-fMRI time series data, offering a more dynamic and flexible representation of brain connectivity. These point clouds allow the method to capture the underlying topological features of functional connectivity across brain regions. Each point represents temporal interactions between regions, making this method particularly sensitive to the non-linear and time-varying nature of brain dynamics. Graph filtration, by contrast, relies on fixed connectivity matrices, which are based on static, pairwise relationships. While this approach captures local connectivity between specific brain regions, it lacks the flexibility to reflect the dynamic changes that occur over time in the brain’s functional network.
In neurodegenerative diseases like MCI, brain networks undergo dynamic reorganization as cognitive decline progresses. The flexibility of Vietoris-Rips filtration allows for the detection of subtle changes in functional connectivity over time, offering a deeper understanding of early disease progression. Capturing these temporal variations is crucial for early diagnosis and intervention in MCI, as it provides a more complete picture of how functional networks are disrupted.

\subsubsection{Inter-ROI Wasserstein distance Vs. lifespan features}
The key distinction between the two approaches lies in the use of inter-ROI Wasserstein distance in Vietoris-Rips filtration. This distance metric quantifies the similarity between persistence diagrams across all ROIs, enabling a more direct and comprehensive comparison between subjects. This offers more global insights into how the overall network topology changes in response to disease progression.
In contrast, graph filtration uses the lifespan of the topological features (connected components in dimension 0 and loops in dimension 1) within the connectivity matrix. This approach focuses on local features, emphasizing persistence within individual brain regions but potentially overlooking global interactions across different parts of the brain.
Neurodegenerative diseases like MCI are characterized by disruptions in inter-regional brain communication and synchronization. The Wasserstein distance, by comparing persistence diagrams across ROIs, better captures the overall dissimilarity in brain network topology between healthy and diseased states. It reflects how disease-related changes impact the global organization of brain networks, including how different regions fall out of synchrony or lose coordination, a hallmark of MCI. In contrast, the lifespan features from graph filtration may miss these more global and inter-regional shifts in connectivity.

\subsubsection{Richer topological features in Vietoris-Rips filtration}
While both graph filtration and Vietoris-Rips capture topological features like connected components (dimension 0) and loops (dimension 1), Vietoris-Rips goes beyond by incorporating dimension 2 (voids), which graph filtration inherently cannot do due to its 1D simplicial complex structure. This additional dimensionality allows for a more comprehensive representation of brain connectivity, capturing complex relationships among brain regions that could reflect higher-order cognitive processes. Study of this higher-order dynamics could be essential in understanding the breakdown of brain networks in MCI. Since cognitive decline affects complex networks, Vietoris-Rips filtration is better suited to reflect the disruptions in these interactions. 
Although individual inter-ROI Wasserstein distance features from each dimension are fed into the CNN, the additional dimensionality $(H_{2})$ provided by Vietoris-Rips filtration enriches the feature set by capturing higher-order interactions, global topological structures, and interdependencies between dimensions. These factors contribute to the improved classification performance, as they offer a more comprehensive representation of the brain's connectivity network, critical for detecting early pathological changes in MCI.
The methodological advantages of Vietoris-Rips filtration, including its capacity to operate with flexible point clouds, the application of inter-ROI Wasserstein distance for global comparisons, and the incorporation of higher-dimensional simplices, hold substantial clinical significance for enhancing our understanding of MCI. The superior performance of Vietoris-Rips filtration in distinguishing MCI sub-types from healthy indivisuals demonstrates its potential to capture complex, dynamic changes in brain connectivity that are crucial for early detection and classification of neurodegenerative diseases.
These factors combine to make Vietoris-Rips filtration a more biologically meaningful and technically robust approach for classifying brain network alterations in MCI, offering deeper insights into the disease’s impact on brain topology and potential for improved diagnosis and intervention strategies.

\subsection{Comparison with state-of-the-art techniques}

To assess the effectiveness of our proposed methodology, we compared the classification results from our approach with those from previous studies. Various methodologies previously have been employed for the automated diagnosis of MCI sub-types and HC, including the sub-network kernel method (\cite{jie_2018_subnetwork}), a multiple-BFN-based 3D CNN framework (\cite{kam_IEEE, kam2018novel}), integrating temporal and spatial properties of network (\cite{jie2018integration}), the Spatial-Temporal convolutional-recurrent neural Network (STNet) (\cite{wang_2020_spatialtemporal}). Although these methods demonstrate commendable performance, our approach of leveraging inter-ROI Wasserstein distance as features for classification, significantly outperforms them. We achieved accuracy of 70.8\% for distinguishing between HC and EMCI, 81\% for HC Vs. LMCI, 77.3\% for differentiating between EMCI and LMCI, and 85.7\% for HC Vs. MCI in the TLSA dataset. Our results specifically highlight the potential of using inter-ROI Wasserstein distance to capture subtle differences in brain connectivity patterns. Table~\ref{table:Comparison} presents a detailed comparison of the classification performance between the proposed methodology and other state-of-the-art approaches, underscoring the robustness and clinical relevance of our findings in the context of MCI diagnosis.

\begin{table}[h]
\begin{center}
\caption{Comparative analysis with recent state-of-the-art techniques that used fMRI data to differentiate MCI sub-types and healthy.}
\vspace{10pt}
\renewcommand{\arraystretch}{1.2}
\label{table:Comparison}
\begin{tabular}{|c|c|c|c|}
    \hline
    \textbf{Reference} & \textbf{Modality} & \textbf{Comparison} & \textbf{Accurcay} \\
    \hline
    \cite{kam2018novel} & fMRI (ADNI) & HC vs EMCI & 74.23\% \\
    \hline
    \cite{jie_2018_subnetwork} & fMRI (ADNI) & EMCI vs LMCI & 74.8\%  \\ \cline{3-4}
    &  & HC vs MCI & 82.6\% \\ 
    \hline   
    \cite{jie2018integration} & fMRI (ADNI) & EMCI vs LMCI & 78.8\% \\
    \hline
    \cite{kam_IEEE} & fMRI (ADNI) & HC vs EMCI & 76.07\% \\
    \hline
    \cite{wang_2020_spatialtemporal} & fMRI (ADNI) & EMCI vs LMCI & 79.36\% \\
    \hline
    \cite{lee_2021_a} & fMRI (ADNI) & HC vs EMCI & 74.42\% \\
    \hline
    \cite{yang2021} & fMRI (ADNI) & HC vs LMCI & 87.23\% \\ \hline
    \cite{bolla2023comparison} & fMRI (ADNI) & HC vs MCI & 90\% \\
    \hline
    \cite{r2024multiscalefmritimeseries} & fMRI (ADNI) & HC vs MCI & 89.47\% \\
    \hline
    \hline
   \multicolumn{4}{|c|}{\textbf{Proposed Methodology (2024)}} \\ 
   \hline
    \multirow{4}{*}{}Graph Filtration & fMRI (ADNI) & HC vs EMCI & 61.8\% \\ \cline{3-4}
    &  & HC vs LMCI & 56.0\% \\ \cline{3-4}
    &  & EMCI vs LMCI & 63.9\% \\ \cline{2-4}
    & In-House TLSA & HC vs MCI &71.4\% \\
    \hline
    \multirow{4}{*}{}Vietoris-Rips Filtration & fMRI (ADNI) & HC vs EMCI & \textbf{70.8\%} \\ \cline{3-4}
    &  & HC vs LMCI & \textbf{81.0\%} \\ \cline{3-4}
    &  & EMCI vs LMCI & \textbf{77.3\%} \\ \cline{2-4}
    & In-House TLSA & HC vs MCI & \textbf{85.7\%} \\
    \hline
    
\end{tabular}
\end{center}
\end{table}

\section{Discussion}
The lack of a standardized clinical test and the absence of a cure for dementia have prompted the exploration of machine learning as a means to identify individuals at risk of developing cognitive impairment, thereby enabling proactive intervention (\cite{stamate_2018_a}). In this context, the application of machine learning techniques has shown significant promise, with studies reporting their ability to accurately differentiate between individuals with early and late stages of mild cognitive impairment, as well as those who are cognitively healthy (\cite{stamate_2018_a, danso_2021_developing, basheera_2019_convolution}). One of the key advantages of using machine learning is its capacity to identify subtle, complex patterns in data that may not be easily discernible to the human eye. This is particularly relevant in the context of neurodegenerative diseases, where the underlying pathological changes often occur years before the onset of clinical symptoms (\cite{danso_2021_developing}).

Numerous studies have attempted to distinguish MCI and its sub-types, often with moderate success in terms of classification accuracy. Unlike traditional methods, our approach introduces an innovative technique grounded in computational topology, specifically utilizing persistent homology from topological data analysis. The proposed study is unique in its application of machine learning combined with persistent homology to explore changes in brain network topology between HC and MCI subtypes (early and late) using fMRI time series data. While most state-of-the-art techniques, such as deep learning and network-based methods, focus on spatial and temporal features or rely on predefined connectivity metrics, our approach captures deeper, more intrinsic properties of brain networks. Persistent homology allows us to identify topological structures, such as connected components, loops, and voids, offering a global view of brain connectivity beyond pairwise interactions. Our study goes beyond conventional network metrics by applying topological data analysis to brain networks, enabling a deeper exploration of the complex geometry and topology of brain function. This approach captures subtle structural differences in brain network topology that are often overlooked by traditional methods. Additionally, Persistent homology is leveraged using two different kinds of filtration techniques: Vietoris-Rips filtration on 3D point cloud and graph filtration on positively correlated network, as constructed from rs-fMRI time series data. This dual approach introduces novelty by offering a more comprehensive understanding of brain network topology. The graph filtration captures local connectivity features, while Vietoris-Rips filtration incorporates higher-dimensional interactions and global topological insights. By comparing these methods, the study highlights how different filtration techniques can reveal unique aspects of brain network reorganization in MCI; setting a new direction for the application of topological data analysis in neurological research.

In case of Vietoris-Rips filtration, the 3D point cloud is generated from 1D fMRI time series using sliding window embedding. Vietoris-Rips filtration is then applied to construct simplicial complexes and compute persistent homology features across dimension-0, 1, and 2. Inter-ROI Wasserstein distances are computed between persistence diagrams for each subject and for each dimension which is then used as features for classification via a CNN. In contrast, for graph filtration, connectivity matrices are constructed from 1D fMRI time series data using partial and marginal correlations. Persistent homology is computed for dimensions 0 and 1, and the top 10 most persistent features are used for classification via a stacked ensemble classifier. Furthermore, inter-subject Wasserstein distances between persistence diagrams as obtained through graph filtration for both homology dimensions are computed to assess statistically significant differences between the study groups.
Though both methods gave decent performance in distinguishing for healthy and MCI, but, classification accuracy obtained using inter-ROI Wasserstein distance between persistent diagrams obtained using Vietoris-Rips filtration as feature significantly outperformed the graph filtration method that used the raw top ten most persistent homology features for classification. This is primarily because of its ability to work with flexible point clouds generated from rs-fMRI time series, rather than being constrained to fixed graph structures. This flexibility allows Vietoris-Rips to capture more detailed topological features across multiple dimensions- dimension-0 (connected components), 1 (loops), and 2 (voids), whereas graph filtration is limited to dimensions 0 and 1. Additionally, the use of inter-ROI Wasserstein distance in Vietoris-Rips filtration offers a more global comparison of brain connectivity patterns, enabling better characterization of the overall network topology. Neurodegenerative diseases like MCI affect large-scale brain networks, disrupting both local and global connectivity. Vietoris-Rips filtration captures these disruptions more effectively, making it a more robust tool for identifying subtle changes in brain network organization that are critical for early detection and classification of MCI subtypes. This enhanced sensitivity to complex network dynamics holds clinical relevance, as it may improve the accuracy of diagnostic models for MCI and other brain disorders.

The present study underscores the clinical potential of persistent homology as a powerful tool for analyzing complex brain network dynamics. By leveraging persistent homology, we were able to capture intricate topological features that traditional methods might overlook, allowing for a more refined differentiation between distinct cognitive states. Our findings demonstrate the effectiveness of this approach in improving the classification accuracy of MCI subtypes, supporting its potential application in clinical diagnostics and decision-making. The novelty of our study lies in its application of persistent homology to identify subtle yet significant topological differences in brain connectivity, which could enhance the precision of MCI diagnoses, aid in tracking disease progression, and inform personalized treatment strategies. Moreover, the success of our model suggests that persistent homology could be extended to a wide range of clinical settings for improving diagnostic accuracy and patient outcomes across various neurological and medical conditions. Our study highlights the potential of integrating persistent homology into clinical workflows, offering the ability to enhance the precision and effectiveness of cognitive assessments. In future, this represents a significant advancement in the field of personalized medicine.

\section{Conclusion}

This study investigated the use of persistent homology techniques, specifically Vietoris-Rips and graph filtration methods, to examine complex brain network dynamics related to MCI. Persistent homology offers a powerful mathematical approach for analyzing topological features within data, and it shows potential in detecting subtle connectivity patterns that may indicate early cognitive decline. Through a comprehensive comparative analysis, our results suggest that Vietoris-Rips filtration may serve as an effective tool for diagnosing and monitoring MCI. This technique provides a refined view of brain connectivity alterations, with implications for advancing research and interventions aimed at early detection and management of cognitive decline.

\bibliographystyle{unsrt}  
\bibliography{PH_MCI_references}

\end{document}